%% file: main_IEEE.tex
\documentclass[conference]{IEEEtran}
\usepackage[dvipsnames]{xcolor}
\usepackage[colorlinks=true,linkcolor=blue,breaklinks=True,citecolor=black,urlcolor=blue]{hyperref}
\usepackage{dingbat}
\usepackage{adjustbox}
\usepackage[linesnumbered,algo2e,ruled]{algorithm2e}
\usepackage{multirow}
\usepackage{booktabs,subcaption,dcolumn}
\usepackage{tikz}
\usepackage{enumitem}
\setlist[itemize]{noitemsep, topsep=0pt}
\usepackage{tabularx}
\usepackage[normalem]{ulem}
\usepackage{fontawesome}
\usepackage{blindtext}
\usepackage{pifont}
\usepackage{svg}
\usepackage{soul}
\usepackage[font=footnotesize]{caption}
\usepackage{amsmath}
\usepackage{amsfonts, amssymb}
\usepackage{colortbl}
\usepackage{tablefootnote}
\usepackage[most]{tcolorbox}
\usepackage[sort&compress, numbers]{natbib}
\usepackage{setspace}

\AtBeginDocument{%
  \providecommand\BibTeX{{%
    \normalfont B\kern-0.5em{\scshape i\kern-0.25em b}\kern-0.8em\TeX}}}

\input{commands.tex}

\begin{document}
    \input{title}
    \input{authors.tex}
    \pagestyle{plain}
    \maketitle
    \input{sections/0-abstract}
    \input{structure}

    \bibliographystyle{IEEEtran}
    {%
        \footnotesize
        \input{bibliography.bbl}

    }

\end{document}

%% file: commands.tex
\newcommand{\cmark}{{\color{ForestGreen}{\ding{51}}\xspace}}
\newcommand{\xmark}{{\color{RubineRed}{\ding{55}}\xspace}}
\newcolumntype{?}{!{\vrule width 1.5pt}}

\newcommand{\fmwrk}[1]{{\small \fontfamily{pcr}\selectfont #1}}
\newcommand{\ftfmwrk}[1]{{\scriptsize \fontfamily{pcr}\selectfont #1}}
\newcommand{\code}[1]{{\small \fontfamily{cmss}\selectfont #1}}

\newcommand{\textbox}[1]{
    \noindent\fbox{%
        \parbox{0.97\columnwidth}{%
            {#1}
        }%
    }
}

\newtcolorbox{cooltextbox}[1][]{%
    colback=black!5,
    colframe=black!5,
    notitle,
    sharp corners,
    borderline west={0pt}{0pt}{red!80!black},
    enhanced,
    breakable,
    left=0pt,
    right=0pt,
    top=0pt,
    bottom=0pt
    }

\newcommand\smamath[1]{{\small $#1$}}

\newcommand\smabb[1]{{\small $\mathbb{#1}$}}

\newcommand\scbb[1]{{\scriptsize $\mathbb{#1}$}}

\newcommand\scmath[1]{{\scriptsize $#1$}}

\newcommand\revision[1]{%
  \bgroup
  \hskip0pt\color{blue!80!black}%
  #1%
  \egroup
}

\newcommand\dataset[1]{{\fontfamily{pcr}\selectfont {\footnotesize #1}}}

%% file: title.tex
\title{Machine Learning in Space: Surveying the Robustness of on-board ML models to Radiation}

%% file: authors.tex
\author{
    \IEEEauthorblockN{{
        Kevin Lange\IEEEauthorrefmark{5},
        Federico Fontana\IEEEauthorrefmark{1},
        Francesco Rossi\IEEEauthorrefmark{1},
        Mattia Varile\IEEEauthorrefmark{1},
        Giovanni Apruzzese\IEEEauthorrefmark{5}}\\
    }
    \IEEEauthorblockA{{ 
        \IEEEauthorrefmark{5}\textit{University of Liechtenstein},
        \IEEEauthorrefmark{1}\textit{AIKO S.r.l.}}\\
    {\small \{kevin.lange, giovanni.apruzzese\}@uni.li\IEEEauthorrefmark{5}, 
    \{federico, francesco, mattia\}@aikospace.com\IEEEauthorrefmark{1}
    }}
}

%% file: sections/0-abstract.tex
\begin{abstract}
Modern spacecraft are increasingly relying on machine learning (ML). However, physical equipment in space is subject to various natural hazards, such as radiation, which may inhibit the correct operation of computing devices. Despite plenty of evidence showing the damage that naturally-induced faults can cause to ML-related hardware, we observe that the effects of radiation on ML models for space applications are not well-studied. This is a problem: without understanding how ML models are affected by these natural phenomena, it is uncertain ``where to start from'' to develop radiation-tolerant ML software. 

As ML researchers, we attempt to tackle this dilemma. By partnering up with space-industry practitioners specialized in ML, we perform a reflective analysis of the state of the art. We provide factual evidence that prior work did not thoroughly examine the impact of natural hazards on ML models meant for spacecraft. Then, through a ``negative result,'' we show that some existing open-source technologies can hardly be used by researchers to study the effects of radiation for some applications of ML in satellites. As a constructive step forward, we perform simple experiments showcasing how to leverage current frameworks to assess the robustness of practical ML models for cloud detection against radiation-induced faults. Our evaluation reveals that not all faults are as devastating as claimed by some prior work. By publicly releasing our resources, we provide a foothold---usable by researchers \textit{without access} to spacecraft---for spearheading development of space-tolerant ML models.
\end{abstract}

%% file: structure.tex
\input{sections/1-introduction}
\input{sections/2-background}
\input{sections/3-related}
\input{sections/4-implementation}
\input{sections/5-results}
\input{sections/6-discussion}
\input{sections/7-conclusions}

%% file: sections/1-introduction.tex
\section{Introduction}
\label{sec:introduction}
\begin{figure}[!htbp]
    \vspace{-2mm}
    \centering
    \includegraphics[width=1\columnwidth]{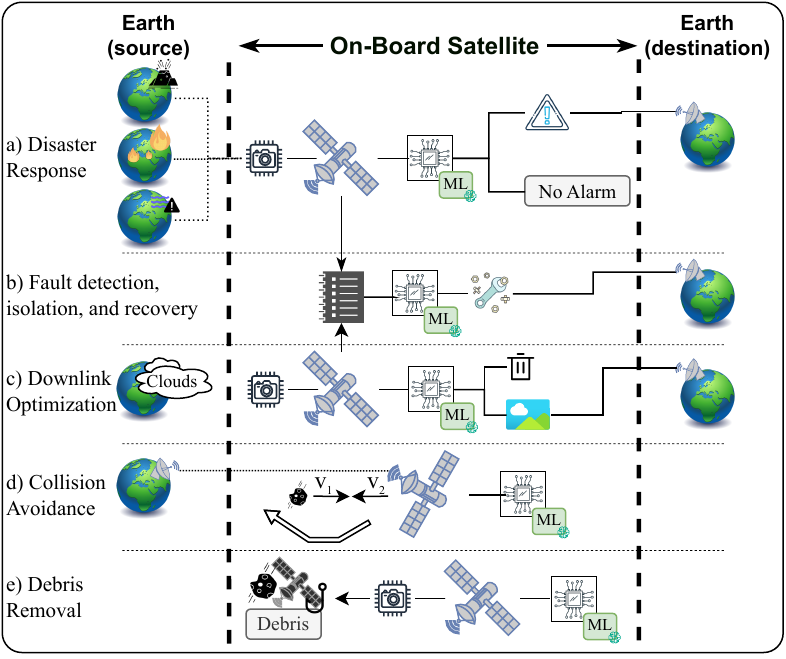}
    \vspace{-0.6cm}
    \caption{\textbf{Applications of ML on-board spacecraft} -- 
    \textmd{\footnotesize In some cases, ML is used to analyse data \textit{of} Earth (taken by the satellite), and then send the results \textit{to} Earth; in other cases, the output of ML is used by the satellite itself.}}
    \label{fig:use_cases}
    \vspace{-2mm}
\end{figure}

\noindent
During the last years, machine learning (ML) solutions for on-board satellite missions have seen a tremendous push from academia, space agencies, as well as industry~\cite{ibm2020}. Indeed, ML can now be used to carry out many space-related tasks~(Fig.~\ref{fig:use_cases}). For instance, thanks to ML, companies in this market can optimize downlink communications, thereby reducing the amount of unusable data and improving efficiency~\cite{bradley2019}. In addition, the increasing interest in space has reduced the time required to launch satellites in orbit---especially for CubeSats~\cite{esa2023CubeSats}, which represent the state of the art of modern spacecraft~\cite{samwel2019Space,rawlins2022death}.

CubeSats are typically equipped with commercial-off-the-shelf (COTS) components (e.g., the NVIDIA Jetson Nano), which can be used to empower ML models~\cite{lofqvist2020}. Aside from being cheaper~\cite{rawlins2022death}, some COTS components can also outperform (e.g., faster processing speed) specialized space-tolerant counterparts~\cite{kothari2020final}. Unfortunately, COTS components have a shorter expected lifetime since they are not designed to withstand the harsh space environment~\cite{cantoro2018}. For instance, compared to radiation hardened components (like the NanoeXplore BRAVE Large or Xilinx Kintex XQRKU060), the Myriad2 can withstand only half of the total ionizing dose~\cite{rapuanoFPGA2021}. Furthermore, Slater et al. found that the NVIDIA Jetson Nano is expected to last at most two years in low Earth orbit~\cite{slater2020}; whereas Rodriguez et al.~\cite{rodriguezFerrandez2022} also found that the NVIDIA Xaview SoC is susceptible to faults induced by natural hazards in space. 

These hazards, such as radiation, are likely to endanger next-generation COTS components even more, due to the manufacturers' interest in \textit{hardware miniaturization}---which exacerbates the vulnerabilities of COTS components to faults such as bit-flips~\cite{rodriguezFerrandez2022}. Worryingly, abundant prior work (e.g.,~\cite{agarwal2022lltfi,rakin2019bit,miller2023space}) showed the disruptive impact that bit-flips can have against some ML models. Therefore, there is a need for \textbf{fault-tolerant software}, which calls for contributions from \textit{various research domains} (e.g., space, but also applied ML)~\cite{crum2022nasa}. 

In this context, we find ourselves in a quandary. On the one hand, a large body of literature (e.g.,~\cite{rakin2019bit}) demonstrated the impact of hardware-related faults on ML models. On the other hand, many papers showed that natural hazards in space can lead to damaged hardware (e.g.,~\cite{slater2020}). However, we ask ourselves: ``what about papers that specifically focus on the impact of \textit{natural hazards on ML models meant for on-board deployment in spacecraft}?'' Indeed, COTS components can empower many computational elements besides ML models; whereas not all the faults that can affect an ML model at the hardware level may be related to natural hazards in space. Hence, before developing fault tolerant software for ML components in space, it is first necessary to determine how much the space environment can affect ML models deployed on-board satellites. We pursue this quest in this paper.

\textsc{\textbf{Contributions.}} We aim to foster development ML methods that are robust to ``natural'' space hazards.
After summarising the applications and problems that entail deployment of ML on satellites (§\ref{sec:background}), we make the following contributions.
\begin{itemize}[leftmargin=0.5cm]
    \item By performing a thorough \textit{literature review}, we show that \textbf{prior work poorly accounted for the effects of natural faults} on ML applications in space (§\ref{sec:sota}); {\scriptsize [major contribution]}
    \item To address this issue, we \textit{analyse open-source toolkits} for carrying out simulations of the effects that radiation can have on ML: through \textit{negative experiments}, we find that \textbf{some current solutions have functional issues} (§\ref{sec:tools}).
    \item To fix this problem, we carry out some \textbf{technical experiments} -- under the guidance of practitioners -- showcasing how to \textit{approximate realistic evaluations}, and the \textit{impact of radiation-induced faults} on ML methods (§\ref{sec:assessment}).
\end{itemize}
Lastly, after outlining implications for related work (§\ref{sec:discussion}), we \textit{publicly release our resources}~\cite{ourRepo}. Besides doing so for scientific reproducibility and transparency, our tools serve to kickstart future experimentation on space-tolerant ML models---without the need to carry out field tests.

\vspace{1mm}

{\setstretch{0.8}
\textbox{{\small \textbf{Remark:} our contributions should \textit{not be taken as a finger-pointing attempt}. Rather, we perform a reflective exercise on the current landscape of papers and technologies for on-board deployment of ML, with the ultimate objective of improving the state of the art.}}}

%% file: sections/2-background.tex
\section{Applications (and Problems) of ML in Space}
\label{sec:background}
\noindent
To setup the stage for our contribution, we summarize the pros-and-cons that entail deployment of ML on-board satellites.

\vspace{1mm}

\begin{figure}[!htbp]
    \centering
    \includegraphics[width=0.8\columnwidth]{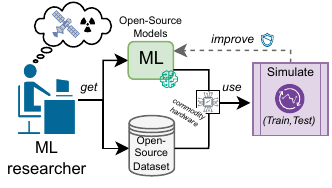}
    \vspace{-3mm}
    \caption{\textbf{Perspective of the ML researcher} --
    \textmd{\footnotesize ML researchers do not have access to spacecraft or to physical equipment that reproduces a space setting. They only rely on open-source tools (models and data) and commodity hardware (e.g., GPUs), but their knowhow can help improve state-of-the-art methods for real-world deployments of ML.}}
    \label{fig:researcher}
    \vspace{-3mm}
\end{figure}

{\setstretch{0.8}
\textbox{{\small \textbf{Goal and Audience.} We seek to build a bridge between two communities: \textit{space researchers}, who have expertise in studying (often via real equipment) the natural phenomena affecting spacecraft; and \textit{ML researchers}, whose proficiency lies in analysing (typically with open-source resources) the ins-and-outs of ML (see Fig.~\ref{fig:researcher})}.}}

\vspace{-3mm}

\subsection{What tasks can be solved by ML (in space)?}
\label{ssec:tasks}
\noindent
Applications of ML for on-board inference in spacecraft involve the processing of information that can be either \textit{used by the satellite itself} or \textit{sent back to Earth}. Indeed, it has been found~\cite{bradley2020} that elaborating the data directly on spacecraft provides many benefits. We identify five exemplary applications (refer to Fig.~\ref{fig:use_cases}) of ML for on-board deployment in Earth-orbit satellites---taken from both research and practice.

\begin{enumerate}[label=\alph*)]
     
    \item \textit{Disaster response}. In these contexts, latency is paramount and can be reduced by transmitting only the most essential information~\cite{izzo2022}. For instance, Ruzicka et al.~\cite{ruzicka2022} used ML to identify the areas affected by earthquakes, fires, floods and other natural disasters. 

    \item \textit{Fault detection, isolation and recovery}. In these cases, the spacecraft uses ML to, e.g., identify faults and issues in its internal pipelines and apply mitigation~\cite{kothari2020final, miralles2021Machine}.
    
    \item \textit{Downlink optimization}. One pioneering example is the $\Phi$-Sat-1 Mission from the European Space Agency in 2020~\cite{giuffrida2021varphi}. Here, ML was used to detect clouds in images~\cite{giuffrida2020CloudScout}, so as to avoid sending useless ``noisy'' images (showing mostly clouds) back to Earth, saving bandwidth.
    
    \item \textit{Collision avoidance}. Due to the growing number of satellites in Earth orbit and the increasing risk of potential collisions between satellites, avoiding such situations becomes increasingly relevant. Gonzalo et al.~\cite{gonzalo2021OnBoard} as well as Bourriez et al.~\cite{bourriez2023spacecraft} showed how to address this problem with ML for on-board computation.
    
    \item \textit{Debris removal}. A fundamental part of mission planning is identifying which targets must be removed, and then pinpoint the most efficient removal order. Notable efforts are the solution by Xu et al.~\cite{xu2023Active} and that by Guthriel~\cite{guthrie2022ImageBased} (which relies on convolutional neural networks).
\end{enumerate}
We also mention some orthogonal applications of ML, e.g., ``federated learning'' approaches~\cite{razmi2022board}; or for missions that go beyond the Earth orbit, such as using ML to pinpoint landing on the Moon~\cite{silvestrini2022optical}, or for managing the autonomy of CubeSats in deep space~\cite{walker2018DeepSpace}. These works are outside our scope.

\subsection{What space-specific problems affect ML (in space)?}
\label{ssec:problems}
\begin{figure}[!htbp]
    \centering
    \includegraphics[width=\columnwidth]{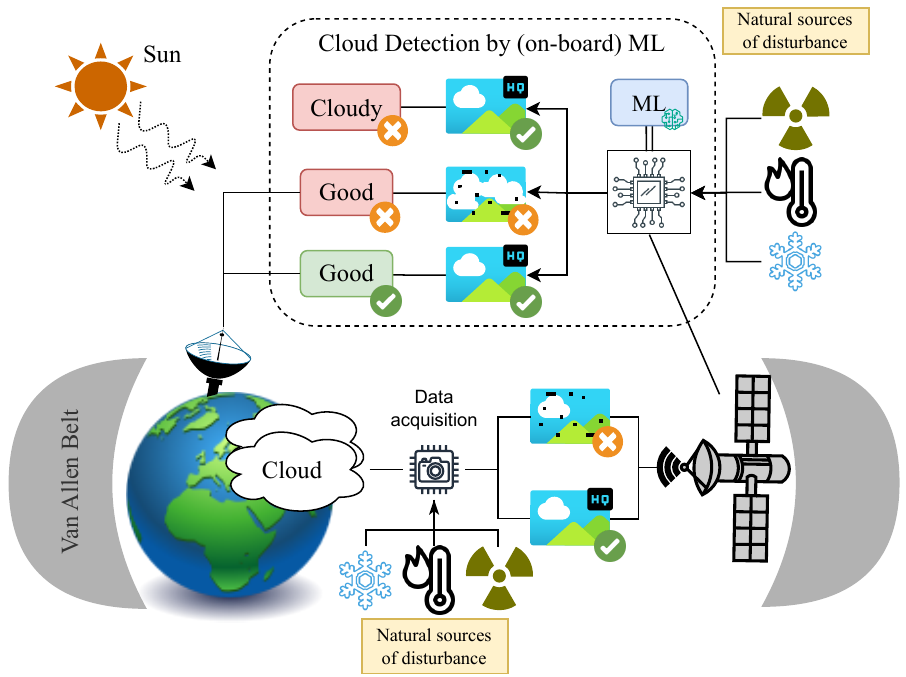}
    \vspace{-0.6cm}
    \caption{\textbf{Using ML for on-board Cloud Detection} --
    \textmd{\footnotesize The satellite acquires data (i.e., images) of Earth; such data may be subject to natural disturbances (e.g., radiation). Then, the captured data is analyzed by an ML model (which may also be subject natural disturbances). The output of this analysis is then sent back to Earth. To optimize downlink communications, ``cloudy'' images (detected via on-board ML) are not transferred. This saves bandwidth.}}
    \label{fig:cloudDetection}
    \vspace{-3mm}
\end{figure}

\noindent
Computing hardware deployed on-board a satellite is exposed to a harsh environment. In particular, two ``natural hazards'' -- which are much more present in space than on Earth -- can interfere with on-board equipment: temperature and radiation (see Fig.~\ref{fig:cloudDetection} for a schema of these hazards for cloud detection).

\subsubsection{\textbf{Temperature}}
\label{sssec:temperature}
In-orbit satellites are exposed to extreme temperatures---both cold and hot. In particular, the surface temperature of low-orbit satellites can vary between \smamath{[-150;+150]} Celsius degrees~\cite{gardo2020System}. Heat sources include: the Sun, the Earth's albedo, the infrared emissions of Earth~\cite{tribak2022Impact}; as well as the heat produced by the satellite itself. 

\subsubsection{\textbf{Radiation}}
\label{sssec:radiation}
Within our solar system we can distinguish three different sources of radiation~\cite{baumann2019}: 
solar radiation, which comes from the Sun in the form of solar wind, solar flares and coronal mass injections~\cite{tekbiyik2022reconfigurable}; 
galactic cosmic rays, coming from outside our solar system, which are remnants particles of galaxies and stars (while most particles are blocked from the heliosphere, others can reach Earth and affect electronic components in satellites~\cite{baumann2019}); and the ``Van Allen Belt'', a toroidal-shaped area of charged particles (whose kinetic energy depends on the Sun's activity~\cite{daglis2001, baumann2019}) which can be found around planets with a magnetic field---such as Earth.

\vspace{-3mm}

The effects of temperature on hardware are well-known in computer science (e.g.,~\cite{platini2018cpu,sarafraz2017convective,handel2008analyzing}), so in the next section we focus on radiation---which is intrinsic of space missions.

\subsection{How can radiation interfere with ML (in-space)?}
\label{ssec:disturbance}
\noindent
Whenever electronic components are exposed to radiation, ionizing particles continuously interact with the device's semiconductors~\cite{cannizzaro2023evaluation}, potentially leading to equipment malfunctions---thereby impacting the ML pipeline in various ways.

\subsubsection{\textbf{Effects on the ML model}}
\label{sssec:effect_ml}
While low-energy ions may not have any effect~\cite{liu2023characterization}, others can cause ``single-event transients'', manifested through memory bit-flips---which may persist until the memory is overwritten (e.g., for the NASA mission Orbview-2, a state recorder had over 200 daily bit-flips~\cite{poivey2003InFlight}). Although bit-flips not necessarily lead to negative consequences, radiation can interfere with an operation causing a ``single event functional interrupt'', which can lead to an incorrect output or a system crash~\cite{loskutov2021investigation}. It is even possible for a component to be permanently damaged (i.e., ``single-event latchups''~\cite{o2015compendium}), e.g., due to high currents which overheat the circuits~\cite{baumann2019}. Altogether, these effects can inhibit the correct operation of the device empowering the ML model---either via hardware- or software-faults~\cite{agarwal2022lltfi,rakin2019bit}. The problem is aggravated for COTS equipment (and its miniaturization): the smaller the manufacturing technology used, the lower the amount of radiation such a technology can tolerate~\cite{rapuanoFPGA2021}.

\subsubsection{\textbf{Effects on Data}}
\label{sssec:effect_data}
\begin{figure}[!htbp]
    \vspace{-3mm}
    \centering
    \includegraphics[width=\columnwidth]{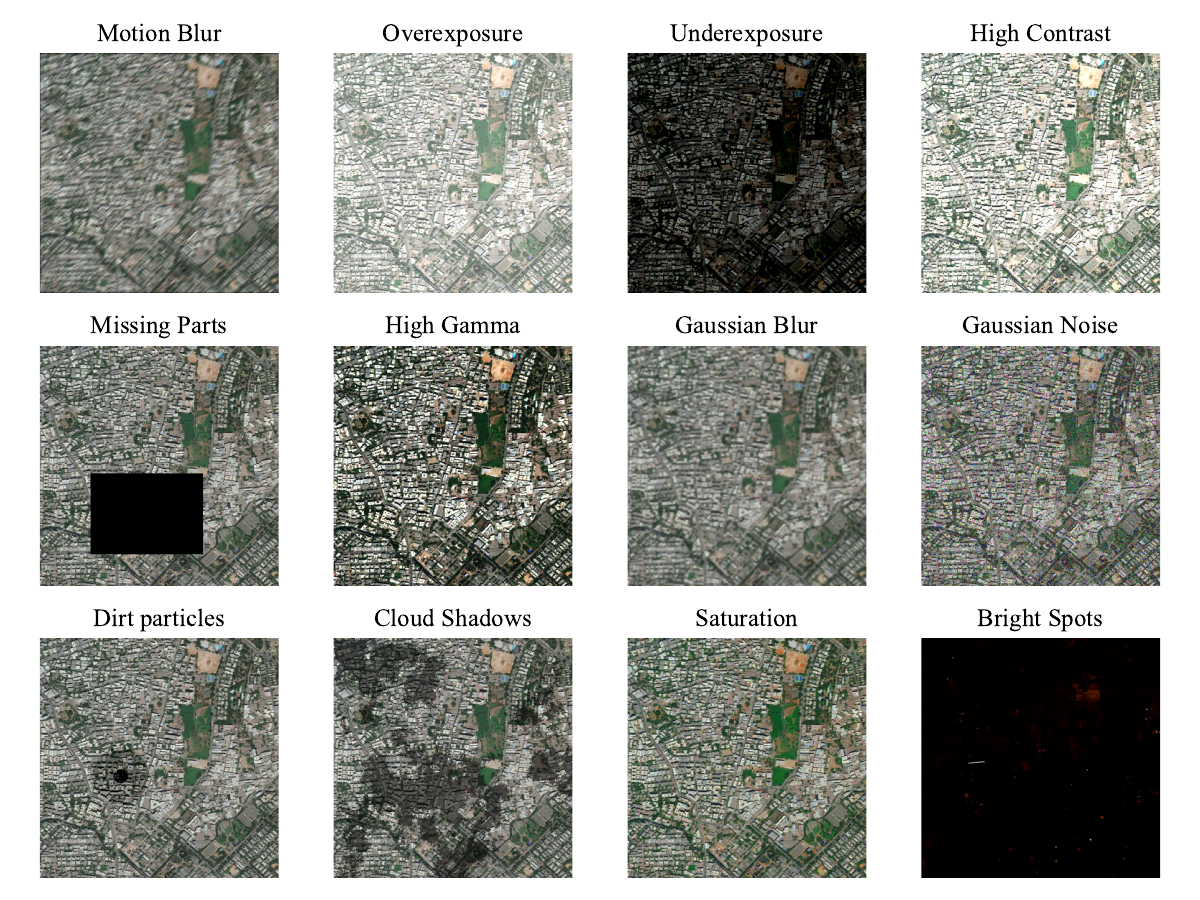}
    \vspace{-0.6cm}
    \caption{\textbf{Possible image disturbances} --
    \textmd{\footnotesize The data (i.e., images) acquired by in-orbit satellites can be perturbed in many ways. Feeding such data to an ML model may ``naturally'' impact its performance. (Own figure, code is at:~\cite{ourRepo})}}
    \label{fig:disturbances}
    \vspace{-3mm}
\end{figure}

Radiation can also affect the data that is meant to be further processed by an ML model. For instance, radiation can lower the picture quality~\cite{virmontois2014radiation} of image sensors reliant on CMOS (which are increasingly used for space applications~\cite{sukhavasi2021cmos}), thereby impacting the accuracy of a cascading ML model. Moreover, sensors' prolonged exposure to radiation increases dark-current in the acquired images~\cite{hopkinson2000Radiation}, potentially leading to permanent damages affecting all taken images~\cite{huber2017Response,cai2020SingleEvent}. Finally, since radiation can cause overheating~\cite{baumann2019}, it can also lead to the complete loss of the camera (such an issue likely affected the Juno mission~\cite{nasa2023}).
We have created a set of images (see Fig.~\ref{fig:disturbances}) showcasing the effects of some disturbances applied to images taken by satellites.

In summary, there is plenty of evidence showcasing the negative effects that such natural hazards can have on components related to ML in space. Thus, we ask ourselves:

\vspace{1mm}

\textbox{\textbf{Research Question \#1:} ``how well does prior research on ML applications in space account for its natural hazards?''}

%% file: sections/3-related.tex
\section{State of the Art (in research)}
\label{sec:sota}
\noindent
We highlighted (§\ref{sec:background}) that abundant prior work showed: {\small \textit{(i)}}~that radiation (and excessive temperature) has negative effects on ML-related components; and {\small \textit{(ii)}}~that such negative effects can impact the performance of ML models. Here, as our first contribution, we scrutinize the extent to which prior research considered the effects of such natural hazards when proposing a ML solutions designed for on-board satellite deployment.\footnote{\textbf{Why is this important?} Among the goals of research is to provide answers to real-world phenomena. However, the research domain does not have always access to real-world equipment---especially for space-related tasks, the costs of such equipment can be prohibitive. Hence, we can expect that most research in this domain carried out their evaluations through simulations. Answering our (first) research question allows one to determine how well the results of prior research approximate those expected from real-world experiments.}

\subsection{Methodology (literature review)}
\label{ssec:method}
\noindent
We carry out a systematic literature review. This entire procedure was done by two authors in Dec. 2023; to ensure consistency, we repeated the same procedure in Feb. 2024.

\textbf{Paper collection.} We mostly rely on the snowball method~\cite{wohlin2014guidelines}. We began our analysis with the papers by Giuffrida et al.~\cite{giuffrida2020CloudScout} and Bruhn et al.~\cite{bruhn2020enabling}, due to their popularity\footnote{As of Feb., 2024,~\cite{giuffrida2020CloudScout} (\cite{bruhn2020enabling}) has $>$100 ($>$55) citations on Google Scholar.}. Next, we provided each of these work as input to ConnectedPapers~\cite{connectedpapers}; and also do a forward/backward snowball search (for both~\cite{giuffrida2020CloudScout,bruhn2020enabling}: altogether, these operations yielded a set of 265 references, which we further expand with the 97 articles listed in OPS-SAT~\cite{opsat}. 

\textbf{Screening.} We manually filter all (362) these documents with the intent of identifying suitable candidates for a deeper analysis. Specifically, we \textit{excluded}: presentations/abstracts, non-peer--reviewed documents, and papers published before 2014 (outdated). Plus, since our focus is on ML, we excluded any paper that did not mention the terms ``machine/deep learning'', ``neural network'', or ``artificial intelligence'' either in the title or in the abstract. Afterwards, we carried out a preliminary inspection of the text, looking for papers that \textit{considered ML applications in space and for on-board satellites} (in Earth orbit---i.e., the ones discussed in §\ref{ssec:tasks}). To this purpose, we excluded any paper that did not mention the term ``on-board'' in the main body; and papers that envision ML applications beyond Earth orbit (e.g.,~\cite{latorre2023transfer}), or do not carry out original ML evaluations (e.g., reviews~\cite{pauly2023survey}). Altogether, these operations led us to a set of  62 papers (published between [2018--2023]), to which we add the work by Haser and Förstner~\cite{haser2022Reliability} which we found thanks to the cooperation with industry practitioners.

\textbf{In-depth analysis.} We manually analyse each of our identified 63 papers by considering six axes: 
{\small \textit{(i)}}~what ML \textit{use-case} is being considered? 
{\small \textit{(ii)}}~does the paper carry out an \textit{on-board evaluation} (i.e., with in-orbit spacecraft) and, if not, is specialized hardware involved? 
{\small \textit{(iii)}}~how many times are the terms ``radiation'' and ``temperature'' \textit{mentioned} in the text?
{\small \textit{(iv)}}~are the effects of ``radiation'' or ``temperature'' considered in the evaluation (either real, or simulated)?
{\small \textit{(v)}}~does the paper \textit{propose} methods to improve the ``robustness'' of ML to natural hazards in space? 
{\small \textit{(vi)}}~is the \textit{code} publicly available? 
Altogether, assessing these axes allow one to provide a comprehensive overview of the state of the art of ML applications in space and w.r.t. their consideration of natural hazards, thereby answering our first RQ. We report the results of our analysis in Table~\ref{tab:sota}. Before analyzing Table~\ref{tab:sota}, however, we reiterate that \textit{our analysis is not a fingerpointing attempt}.

\input{sections/tab_related}

\subsection{Findings (and interpretations)}
\label{ssec:findings}
\noindent
From Table~\ref{tab:sota}, three findings are apparent. Out of 63 papers:
\begin{itemize}
    \item Only 3 papers (\smamath{5\%}) propose solutions to make ML methods more ``robust'' against natural space hazards.
    \item 29 papers (\smamath{46\%}) \textit{never} mention ``radiation'', and 41 (\smamath{65\%}) papers \textit{never} mention ``temperature''
    \begin{itemize}
        \item and 61 (\smamath{97\%}) papers mention ``radiation'' or ``temperature'' less than 6 times.
    \end{itemize}
    \item Only 7 papers (\smamath{11\%}) release their implementation.
\end{itemize}
Let us now analyze these results at a low-level, focusing on ``where'' each paper carries out its evaluation.

\textbf{In-orbit evaluations} (dark-gray rows). Seven (\smamath{11\%}) papers perform evaluations on in-orbit spacecraft. Their results undeniably represent the real world, as they are obviously affected by natural hazards. For instance, Giuffrida et al.~\cite{giuffrida2021varphi}, consider radiation-hardened FPGA, and conclude that such components are ``robust''; however, such a statement is derived by simply observing that the resulting performance of the ML models is high despite the exposure to radiation. As a matter of fact, these works \textit{do not underscore the impact of natural hazards} on the corresponding results (i.e., there is no ``hazard-free'' baseline that can be used to study the effects of such hazards). Indeed, these papers \textit{mention} the term ``radiation'' (typically to account for limitations of the results which are likely affected by radiation), but do not study this phenomenon. 

\textbf{Hardware-reliant simulations} (light-gray rows). The majority (30, \smamath{47\%}) of papers employ space-related hardware in their assessments. Among these, the most ``accessible'' work to ML researchers is the one by Bruhn et al.~\cite{bruhn2020enabling} which focuses on GPU testing, but the datasets considered (i.e., \dataset{MNIST}) have little relevance with space settings. Some simulations carried out on space-compliant hardware typically envisioned in space settings do not consider radiation-tolerant COTS components. For instance, Rosso et al.~\cite{delrosso2021board} experiment on the Myriad2, which is weak to radiation (as shown in~\cite{rapuanoFPGA2021}); the opposite is done, e.g., in Pitonak et al.~\cite{pitonak2022cloudsatnet}, whose experiments are run on radiation-hardened FPGA---but in both cases, there is no assessment of the actual effects of radiation. Notably, Azami et al.~\cite{azami2022earth} carry out a radiation test for six hours on a Raspberry Pi, showing that a single event latchup occurred after 5hours which increased the power consumption---but there was no measure of how it affected the performance of the ML model. A similar (and far more realistic) experiment was done by Buckley et al.~\cite{buckley2022radiation}---but even here, the focus was more on the response of the Myriad2 rather than on the performance of the ML model. Nevertheless, \textit{such tests cannot be replicated by a ML researcher} without specialized equipment (and the code of~\cite{azami2022earth,buckley2022radiation,bruhn2020enabling} is not public). Notably, however, two papers\footnote{Such data was obtained with drones capturing images from Earth. We considered these as ``hardware'' and not ``real-space'' because the altitude of the drone was not high enough to be exposed to the natural hazards of space.} by Del Rosso et al~\cite{delrosso2021board, delrosso2022demo} release their data.

\textbf{Software-based experiments} (white rows).
A total of 26 (\smamath{41\%}) papers carry out evaluations at the software-level (requiring only, e.g., a GPU).  However, most of such papers simply aim at improving the baseline accuracy of ML techniques \textit{and do not account for radiation}. Remarkably, Garrett and George~\cite{garrett2021improving} seek to improve the robustness of tensorflow-based ML models to radiation. However, despite mentioning various ML applications (e.g., ``debris tracking''), the experiments were carried out on common benchmark datasets (i.e., \dataset{MNIST}), which are not representative of a realistic space environment---a limitation affecting also~\cite{haser2022Reliability}. Furthermore, the code of~\cite{garrett2021improving,haser2022Reliability,shi2023automated} is also not available---which is, unfortunately, a common occurrence in our analysis.\footnote{\textbf{Disclaimer:} We stress that many works do provide extensive details for reproducibility. For instance, Fratini et al.~\cite{fratini2022board} leverage the NanoSat framework and abundant details are provided in the paper. Hence, lack of source code does not prevent scientific reproducibiilty---despite inevitably delaying further developments (at least from the perspective of ML researchers).}

\subsection{Consequences (and the way forward)}
\label{ssec:consequences}
\noindent
Our analysis reveals a fundamental problem: not only {\small \textit{(i)}}~most prior research \textit{does not account} for the effects of natural hazards on ML applications in space; but also {\small \textit{(ii)}}~those papers for which the effects are implicitly considered (e.g.,~\cite{bruhn2020enabling,giuffrida2020CloudScout}) \textit{do not allow one} to determine their impact on the ML's output; and {\small \textit{(iii)}}~\textit{few works openly release} their implementation. Put differently, from a research perspective, it is currently unknown how to approximate (and, hence, reproduce) the effects that natural hazards have on real-world deployments of ML for on-board satellites. This is the source of our quandary.

\vspace{-2mm}

\begin{cooltextbox}
\textsc{\textbf{Takeaways.}} Only few works considered the effects that space-natural hazards can have on ML models,\footnote{Even a recent ``critical analysis'' poorly accounts for radiation~\cite{miralles2023critical}.} and few papers share their resources---hindering reproducibility.

\vspace{-2mm}
\rule{0.2\textwidth}{0.1pt} 

\vspace{-3mm}

\end{cooltextbox}

If one can truly measure the effects of such hazards, one can also determine which solution can be used to mitigate their impact. Indeed, there exist many techniques that specifically focus on improving the robustness of ML models against bit-flips (e.g., selective hardening~\cite{libano2018selective}). However, all such techniques have been proposed for tasks that do not strictly pertain to ML applications in space. For instance,~\cite{adam2021ASelective} focuses on healthcare; whereas~\cite{ibrahim2020Soft} considers generic object recognition (not in space); furthermore, the experiments in~\cite{libano2018selective} are carried out on \dataset{MNIST}. Hence, we ask ourselves:

\vspace{1mm}
{\setstretch{0.9}
\textbox{\textbf{Research Question \#2:} ``can an ML researcher -- without access to specialized equipment -- reproduce the effects of natural hazards, and specifically of \textit{radiation}, on ML applications in space by leveraging current technologies?''}
}

%% file: sections/tab_related.tex
\newcommand{\real}{\rowcolor{gray!40}}
\newcommand{\hw}{\rowcolor{gray!15}}
\newcommand{\sw}{}

\begin{table}[!htbp]
    \centering
    \caption{\textbf{Papers on ML applications onboard spacecraft} --
    \textmd{\footnotesize We used the snowball method starting from~\cite{bruhn2020enabling,giuffrida2020CloudScout} and adding~\cite{opsat}: from 362 documents, we distill 63 papers. Use-cases are taken from §\ref{ssec:tasks} {\scriptsize (ColAvo=Collision Avoidance, DisRes=Disaster Response, DebRem=Debris Removal, DlkOpt=Downlink Optimization, FDIR=Fault Detection Isolation Recovery}); parentheses denote potential misalignment between testbed and use-case. Row color denotes the experimental settings (white=software, light-gray=space-related hardware, dark-gray=in-orbit spacecraft). In ``Rad?'' and ``Temp?'', we report the number of times ``radiation'' and ``temperature'' occur in the text; and a \cmark\ (\xmark) denotes whether they affect the results (or not). For ``robust'', the paper had to \textit{evaluate} some hardening/robustness method (not just ``mention'') to get a \cmark. For papers that release ``Code'', the \cmark\ points to the public repository; {\tiny \faFile} is for data only.}}
    \label{tab:sota}
    \vspace{-2mm}
    \resizebox{1\columnwidth}{!}{
        \begin{tabular}{c|c|c|c|c|c|c}
            \toprule
            \textbf{Paper} & \multirow{2}{*}{\textbf{Year}} &\multirow{2}{*}{\textbf{Use-Case}} & \multirow{2}{*}{\textbf{Rad?}} & \multirow{2}{*}{\textbf{Temp?}} & \multirow{2}{*}{\textbf{Robust?}} & \multirow{2}{*}{\textbf{Code?}}\\
            \textbf{(1st author)} & & &  &  &  &  \\

            \midrule

            \sw{} Li~\cite{li2018onboard} & 2018 & DlkOpt & 5 (\xmark) & 5 (\xmark) &   & \\

            \hw{} Pastena~\cite{pastena2019esa} & 2019 & DlkOpt & 0 (\xmark) & 0 (\xmark) &   &  \\
            \hw{} Esposito~\cite{esposito2019orbit} & 2019 & DisRes & 0 (\xmark) & 4 (\xmark) &  & \\

            \hw{}Lemaire~\cite{lemaire2020fpga} & 2020 & DlkOpt & 0 (\xmark) & 0 (\xmark) & & \\
            \hw{}  Furano~\cite{furano2020towards} & 2020 & DlkOpt & 36 (\cmark) & 4 (\xmark) &   & \\
            \hw{}  Kothari~\cite{kothari2020final} & 2020 & DlkOpt & 6 (\cmark) & 1 (\xmark) &   & \\
            \real{} Denby~\cite{denby2020orbital}  & 2020 & DlkOpt & 2 (\cmark) & 4 (\cmark) &   & \href{https://github.com/CMUAbstract/oec-asplos20-artifact}{\cmark} \\
            \sw{} Maskey~\cite{maskey2020cubesatnet} & 2020 & DlkOpt & 0 (\xmark) & 0 (\xmark)  &  & \\
            \real{} Giuffrida \cite{giuffrida2020CloudScout} & 2020 & DlkOpt & 7 (\cmark) & 0 (\cmark) &  & \\
            \hw{} Bruhn~\cite{bruhn2020enabling} & 2020 & DlkOpt & 34 (\cmark) & 1 (\cmark) & \cmark & \\
            \hw{} Reiter~\cite{reiter2020fpga} & 2020 & DlkOpt & 7 (\xmark) & 2 (\xmark) &  & \\
            
            \hw{} Mateo-Gracìa~\cite{mateo2021towards} & 2021 & DisRes & 0 (\xmark) & 0 (\xmark) & & \href{https://gitlab.com/frontierdevelopmentlab/disaster-prevention/cubesatfloods}{\cmark} \\
            \real{} Giuffrida~\cite{giuffrida2021varphi} &  2021 & DlkOpt & 9 (\cmark) & 1 (\cmark) & & \\
            \hw{} Rapuano~\cite{rapuano2021fpga} & 2021 & DlkOpt & 13 (\xmark) & 1 (\xmark) & & \\
            \hw{} Del Rosso~\cite{delrosso2021board} & 2021 & DisRes & 0 (\xmark) & 3 (\xmark) & & \href{https://github.com/alessandrosebastianelli/OnBoardVolcanicEruptionDetection}{{\footnotesize \faFile}} \\
            \hw{}Kucik~\cite{kucik2021investigating} & 2021 & DisRes & 1 (\xmark) & 0 (\xmark) & & \href{https://github.com/AndrzejKucik/SNN4Space}{\cmark} \\
            \hw{}Diana~\cite{diana2021oil} & 2021 & DlkOpt & 14 (\xmark) & 0 (\xmark) & & \\
            \sw{}Ferrari~\cite{ferrari2021integrating} & 2021 & DlkOpt & 0 (\xmark) & 0 (\xmark) & & \\
            \hw{}Pacini~\cite{pacini2021multi} & 2021 & DlkOpt & 1 (\xmark) & 0 (\xmark) & & \\
            \hw{}Leong~\cite{leong2021image} & 2021 & DlkOpt & 0 (\xmark) & 0 (\xmark) & & \\
            \sw{}Fernando~\cite{fernando2021towards} & 2021 & DisRes & 0 (\xmark) & 0 (\xmark) & & \\
            \sw{}Garrett~\cite{garrett2021improving} & 2021 & (DebRem) & 7 (\cmark) & 0 (\xmark) & \cmark &  \\

            \sw{} Haser~\cite{haser2022Reliability} & 2022 & (FDIR) & 10 (\cmark) & 0 (\xmark) & & \\
            \sw{}Farr~\cite{farr2022self} & 2022 & DlkOpt & 0 (\xmark) & 0 (\xmark) & & \\
            \sw{}Růžička~\cite{ruuvzivcka2022ravaen} & 2022 & DisRes & 0 (\xmark) & 0 (\xmark) & & \href{https://github.com/spaceml-org/RaVAEn}{\cmark} \\
            \hw{}Azami~\cite{azami2022earth} & 2022 & DisRes & 10 (\cmark) & 12 (\xmark) & & \\ 
            \sw{}Luo~\cite{luo2022lwcdnet} & 2022 & DlkOpt & 0 (\xmark) & 2 (\xmark) & & \\
            \real{}Labrèche~\cite{labreche2022ops} & 2022 & {\scriptsize DlkOpt/FDIR} & 2 (\cmark) & 0 (\cmark) & & \href{https://github.com/georgeslabreche/opssat-smartcam/tree/v2.1.2}{\cmark} \\
            \hw{}Spiller~\cite{spiller2022wildfire} & 2022 & DisRes & 1 (\xmark) & 2 (\xmark) & & \\
            \sw{}Luo~\cite{luo2022ecdnet} & 2022 & DlkOpt & 0 (\xmark) & 1 (\xmark) & & \\
            \sw{}Salazar~\cite{salazar2022cloud} & 2022 & DlkOpt & 0 (\xmark) & 0 (\xmark) & & \\
            \hw{}Pitonak~\cite{pitonak2022cloudsatnet} & 2022 & DlkOpt & 3 (\xmark) & 1 (\xmark) & & \\
            \sw{}Guerrisi~\cite{guerrisi2022satellite} & 2022 & DlkOpt & 0 (\xmark) & 0 (\xmark) & & \\
            \hw{}Jeon~\cite{jeon2022channel} & 2022 & DlkOpt & 0 (\xmark) & 0 (\xmark) & & \\
            \hw{}Zeleke~\cite{zeleke2023new} & 2022 & DlkOpt & 1 (\xmark) & 0 (\xmark) & & \\
            \hw{}Del Rosso~\cite{delrosso2022demo} & 2022 & DisRes & 0 (\xmark) & 1 (\xmark) & & \href{https://github.com/alessandrosebastianelli/OnBoardVolcanicEruptionDetection}{{\footnotesize \faFile}}\\
            \hw{}Murphy~\cite{murphy2022developing} & 2022 & FDIR & 1 (\xmark) & 2 (\xmark) & & \\
            \hw{}Spiller~\cite{spiller2022hardware} & 2022 & DisRes & 0 (\xmark) & 4 (\xmark) & & \\
            \hw{}Buckley~\cite{buckley2022radiation} & 2022 & DlkOpt & 31 (\cmark) & 8 (\cmark) & & \\
            \real{}Mateo-Gracìa~\cite{mateogarcia2023orbit} & 2023 & DisRes & 1 (\cmark) & 0 (\cmark) & & \href{https://github.com/spaceml-org/ml4floods}{\cmark} \\
            \sw{}Labrèche~\cite{labreche2022artificial} & 2022 & DlkOpt & 0 (\xmark) & 0 (\xmark) & & \href{https://github.com/georgeslabreche/dkm/tree/opssat}{\cmark}\\
            \real{}Abderrahmane~\cite{abderrahmane2022spleat} & 2022 & DlkOpt & 0 (\cmark) & 0 (\cmark) & & \\
            \real{}Kacker~\cite{kacker2022machine} & 2022 & DlkOpt & 3 (\cmark) & 7 (\cmark) & & \\
            \hw{}Mladenov~\cite{mladenov2022augmenting} & 2022 & DlkOpt & 1 (\xmark) & 0 (\xmark) &  & \\
            \hw{}Kervennic~\cite{kervennic2022deployment} & 2022 & DlkOpt & 0 (\xmark) & 0 (\xmark) & & \\
            \sw{}Fratini~\cite{fratini2022board} & 2022 & DlkOpt & 0 (\xmark) & 0 (\xmark) & & \\

            \sw{}Gu~\cite{gu2023learning} & 2023 & DlkOpt & 0 (\xmark) &  0 (\xmark) & & \\
            \sw{}Guerrisi~\cite{guerrisi2023artificial} & 2023 & DlkOpt & 1 (\xmark) & 0 (\xmark) & & \\
            \sw{}Cascelli~\cite{cascelli2023use} & 2023 & DlkOpt & 1 (\xmark) & 0 (\xmark) & & \\
            \hw{}Coca~\cite{coca2023fpga} & 2023 & DisRes & 4 (\xmark) & 1 (\xmark) &  & \\
            \sw{}Shi~\cite{shi2023automated} & 2023 & DlkOpt & 1 (\cmark) & 0 (\xmark) & \cmark & \\
            \sw{}Serief~\cite{serief2023deep} & 2023 & DlkOpt & 2 (\xmark) & 0 (\xmark) & & \\
            \hw{}Kadway~\cite{kadway2023low} & 2023 & DlkOpt & 0 (\xmark) & 0 (\xmark) & & \\
            \sw{}Ferrante~\cite{ferrante2022fault} & 2023 & FDIR & 1 (\xmark) & 16 (\xmark) & & \\
            \hw{}Carbone~\cite{carbone2023comparison} & 2023 & DlkOpt & 1 (\xmark) & 0 (\xmark) & & \\
            \hw{}Ciardi~\cite{ciardi2022gpu} & 2023 & FDIR & 7 (\xmark) & 0 (\xmark) & & \\
            \sw{}Deticio~\cite{deticio2023improving} & 2023 & DlkOpt & 0 (\xmark) & 0 (\xmark) & & \\
            \sw{}Deticio~\cite{deticio2023application} & 2023 & DlkOpt & 0 (\xmark) & 0 (\xmark) & & \\
            \hw{}Leon~\cite{leon2023accelerating} & 2023 & ColAvo & 3 (\xmark) & 0 (\xmark) & & \\
            \sw{}Fernando~\cite{fernando2023towards} & 2023 & DlkOpt & 0 (\xmark) & 0 (\xmark) & & \\
            \sw{}Murphy~\cite{murphy2023overview} & 2023 & FDIR & 0 (\xmark) & 0 (\xmark) & & \\
            \sw{}Nalepa~\cite{nalepa2023look} & 2023 & FDIR & 0 (\xmark) & 0 (\xmark) & & \\
            \sw{}Bourriez~\cite{bourriez2023spacecraft} & 2023 & ColAvo & 1 (\xmark) & 0 (\xmark) & & \\
            \bottomrule
        \end{tabular}
    }
\end{table}

%% file: sections/4-implementation.tex
\section{Tools for reproducing Radiation's effects}
\label{sec:tools}
\noindent
We now analyse the publicly available toolkits that a researcher can use to replicate a realistic environment. In particular, we consider the task of downlink optimization via \textit{cloud detection}, and we focus on the impact that \textit{radiation} can have on the corresponding ML pipeline either by bit-flips or by image distortion.\footnote{\textbf{Why do we focus on this?} The problem of on-board cloud detection via ML is \textit{popular} in prior research (among the 42 papers within the domain of downlink optimization in Table~\ref{tab:sota}, 20 consider it); furthermore, we consider radiation-induced bit-flips because we do not want to \textit{physically damage} our own (real) hardware---plus, single events latchups, while possible, are rare in space missions~\cite{rapuanoFPGA2021} (confirmed by our practitioners); nonetheless, since this task entails analysing images, we can reproduce the effects of (partially) broken sensors on the acquired data. We stress that \textit{our analysis is just a first step}, and we do not claim generality (albeit our findings are applicable also to other use-cases, e.g., disaster response).} 
Importantly, for this analysis \textbf{we rely on the know-how of} AIKO S.r.l., an European company specialized in the development (and deployment) of ML in spacecraft.

\subsection{Resources for replicating on-board cloud detection via ML}
\label{ssec:resources}
\noindent
Assume that we want to carry out realistic experiments on ML applications in space, focusing on cloud detection (see Fig.~\ref{fig:cloudDetection}), but without having access to a real spacecraft or physical equipment (e.g.,~\cite{miller2023space, bruhn2020enabling, sabogal2019recon}). 
To achieve such a goal, we must: {\small \textit{(i)}}~get an ML model---potentially by drawing from state-of-the-art methods; {\small \textit{(ii)}}~train it over one dataset---potentially by ensuring that it achieves state-of-the-art performance; {\small \textit{(iii)}}~and then find a way to replicate the effects of radiation by either {\small \textit{(a)}}~manipulating the input test-data, or by {\small \textit{(b)}}~introducing bit-flips that affect the processing of the ML model (see Fig.~\ref{fig:researcher}).

\textbf{Space-specific resources.}
For the first two steps, we can certainly rely on well-known tools within the ML community: e.g., TensorFlow/PyTorch frameworks, or the \dataset{ImageNet} dataset.\footnote{One can also use the code developed by prior research. Unfortunately, as shown in Table~\ref{tab:sota}, few papers publicly release their artifacts.} However, we highlight a repository~\cite{satelliteGitHub} (which has over 7k stars as of January 2024) that provides plenty of resources for space-related experiments (such as the \dataset{95-Cloud} dataset~\cite{mohajerani2020cloud}). Unfortunately, after extensively analysing this repository we found that \textit{it provides no suggestion/tool to replicate the effects of radiation} (and not even of extreme temperatures) on ML components. Even recently proposed space-specific ``testing labs/implementations'' do not envision either ML (e.g.~\cite{timpe2023application}) or the effects of natural hazards (e.g.,~\cite{costin2023towards}). We even inquired practitioners if they were aware of any open-source and space-specific platform that provided such a functionality, and they were not aware of any such tool.

\textbf{Tools for fault-injection.}
According to practitioners, the only viable way to replicate the effects of radiation (and of natural hazards in general) is to: {\small \textit{(i)}}~``assume'' that a given component is exposed to non-negligible radiation (see §\ref{ssec:disturbance}), {\small \textit{(ii)}}~``guess'' the effects of such radiation on the given component; and then {\small \textit{(iii)}}~manually ``inject'' the corresponding fault/disturbance. For this purpose, we found\footnote{We perform this search by looking at the frameworks mentioned in prior work, as well as by manually searching over the repositories linked in~\cite{satelliteGitHub}.} three notable tools that can be leveraged for on-board cloud detection. 
\begin{itemize}
    \item NVIDIA Binary Instrumentation Tool for Fault Injection (\fmwrk{NVBitFI})~\cite{nvbitfti} (paper:~\cite{tsai2021nvbitfi}), useful to \textit{emulate} faults at the hardware level (e.g., on the GPU accelerator);

    \item Low-Level Tensor Fault Injector (\fmwrk{LLTFI})~\cite{lltfi} (paper:~\cite{agarwal2022lltfi}, useful to \textit{simulate} faults at the software level (e.g., by operating on the ML model during its analysis).

    \item \fmwrk{Kornia}~\cite{kornia} (paper:~\cite{eriba2019kornia}), which is a library for image augmentation and hence useful to inject disturbances in images (we used Kornia to make Fig.~\ref{fig:disturbances}).
\end{itemize}
We also mention \fmwrk{PyTorchFI}~\cite{mahmoud2020pytorchfi} and \fmwrk{TensorFI}~\cite{chen2020tensorfi}, which are specifically tailored for PyTorch and TensorFlow. However, after inquiring practitioners, they discouraged from using these tools since they operate at a much higher level than, e.g., \fmwrk{LLFTI}; furthermore, \fmwrk{TensorFI}'s (whose last commit in its repo dates back to 2021) does not support Python3 and TensorFlow2---which are the state of the art for ML.

\subsection{Analysis of \fmwrk{NVBitFI} and \fmwrk{LLTFI} (negative result)}
\label{ssec:negative}
\noindent
We now seek to use the identified tools to carry out a proof-of-concept evaluation of on-board ML for cloud detection.
Ideally, faults injected at the hardware level are more realistic, since cosmic radiation affects physical equipment. Hence, a researcher interested in replicating a real setting for on-board cloud detection via ML would first opt for \fmwrk{NVBitFI}, and only consider \fmwrk{LLTFI} as a last resort.

\subsubsection{\textbf{\fmwrk{NVBitFI} (scarce documentation)}}
\label{sssec:nvbitfi}
After reviewing the repository of \fmwrk{NVBitFI}~\cite{nvbitfti} (and its research paper~\cite{tsai2021nvbitfi}) we found that, in its documentation, \textit{there are no instructions on how to use} \fmwrk{NVBitFI} for ML-specific experiments. We tried to infer some low-level details from papers that claim to have used \fmwrk{NVBitFI} for ML evaluations, but we could not find any meaningful instruction: most papers (e.g.,~\cite{garrett2021improving}) do not release the source-code. The only relevant effort is the paper by Dos Santos et al.~\cite{dos2022characterizing}, providing a forked version of the \fmwrk{NVBitFI} repository which includes some ``test-apps'' that also contain ML models: however, these ML models are not functional (they are either old, or written in C) for space-related applications, and there is scarce documentation on how to setup the environment to develop alternative ML models. We hence posit that, although \fmwrk{NVBitFI} can theoretically be used for ML experiments, its current implementation can hardly support space-specific evaluations. Even by inquiring practitioners, we were told that they did not know how to setup a space-compliant and ML-ready testbed through \fmwrk{NVBitFI}.

\subsubsection{\textbf{\fmwrk{LLTFI} (problems and errors)}}
\label{sssec:lltfi}
\begin{figure}[!htbp]
    \centering
    \includegraphics[width=\columnwidth]{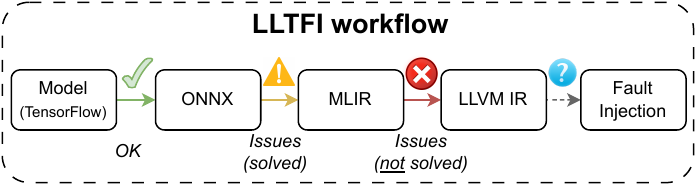}
    \vspace{-0.6cm}
    \caption{\textbf{Our negative experiment} --
    \textmd{\footnotesize We followed the guidelines provided by the developers of {\scriptsize \fontfamily{pcr}\selectfont LLTFI}. We could not finish the workflow due to a fatal error for which we found no workaround (even after consultation with practitioners).}}
    \label{fig:negative}
    \vspace{-3mm}
\end{figure}

We then turn our attention to \fmwrk{LLTFI}, which is specifically designed for ML experimentation. The documentation in the \fmwrk{LLTFI} repository~\cite{lltfi} provides detailed instructions, which we rigorously follow. Specifically, after installing the \fmwrk{LLTFI} framework, we use TensorFlow to train an U-Net (an image-segmentation ML model, representing the state-of-the-art for cloud detection~\cite{mohajerani2019cloud}) on the \dataset{95-Cloud} dataset~\cite{mohajerani2020cloud} (we apply an 80:20 train-test split). After having trained the model (and verifying that its performance aligns with the state of the art on this dataset), we follow the guidelines of \fmwrk{LLTFI} and successfully convert our model in ONNX format. However, from now on we began to encounter ``unexpected'' issues (see Fig.~\ref{fig:negative}).

\vspace{-3mm}

\begin{itemize}[leftmargin=0.5cm]
    \item \textit{ONNX$\rightarrow$MLIR}: before injecting the faults, \fmwrk{LLTFI} requires the model (in ONNX format) to be converted into its MLIR representation, which can be done with an ONNX-MLIR conversion-tool linked in the repository~\cite{onnxmlirlltfi}. We were not able to complete this procedure due to an error (``\code{unhandled option in ConvTranspose}''), which we hypothesized was related to the Conv2DTranspose layer. We tried changing this layer to an UpSampling2D (envisioned in the U-Net used, e.g., in~\cite{guo2021cloudet}), and we also were not successful (``\code{not implemented yet}'', suggesting that this layer was not supported). We found a workaround by using the most-recent version of the ONNX-MLIR tool~\cite{onnxmlir}---\textit{not provided in the} \fmwrk{LLTFI} repository.
    \item \textit{MLIR$\rightarrow$LLVM-IR}: \fmwrk{LLTFI} injects the faults at a low level, which requires the MLIR version of the model to be further converted into LLVM-IR. We were not able to complete this procedure with the instructions in \fmwrk{LLTFI}'s repository: when we run the commands in the docs, we encounter an error (``\code{operation being parsed with an unregistered dialect}''), and upon following the provided suggestion we encounter another error (``\code{custom op 'memref.dim' is unknown}'') which we could not troubleshoot.
\end{itemize}

We searched in the discussion section of the \fmwrk{LLTFI} repository, but we could not find a solution. After consultation with practitioners,\footnote{\textbf{Practitioners' Feedback:} we inquired the practitioners opinion on (potentially) using \ftfmwrk{LLTFI} and \ftfmwrk{NVBitFI} for their simulations meant to improve the robustness of ``radiation-hardened'' ML prototypes. Accordingly, \ftfmwrk{NVBitFI} is not ideal, since its workflow poses a high overhead. Indeed, companies typically convert ML models in ONNX for their applications, hence \ftfmwrk{LLTFI} represents a more suitable solution---if it worked!}  even they were not able to resolve the issue. This ``failed experiment'' is provided in our repository~\cite{ourRepo}.

\subsection{Reflections and remediations}
\label{ssec:reflections}
\noindent
Our analysis leads us to derive two takeaways---but we reiterate that we are \textit{not pointing-the-finger} (see disclaimer §\ref{sec:introduction}):

\begin{cooltextbox}
\textsc{\textbf{Takeaways.}} (1)~There is a lack of open-source resources that facilitate realistic experiments for on-board ML. (2)~The few existing toolkits do not allow to setup a testbed that supports state-of-the-art ML methods for cloud detection.

\end{cooltextbox}

\noindent
Inspired by these (negative) findings -- which lead to an unsatisfactory answer to our RQ\#2 -- we posit that \textit{it is still possible} to assess the effects of (radiation-induced) faults on a representative ML pipeline for cloud detection. This requires to introduce such ``bugs'' by manually tampering with the (trained) ML model---i.e., by directly manipulating its weights (thereby simulating a bit-flip~\cite{rakin2019bit}). We hence ask ourselves:

\vspace{1mm}

\textbox{\textbf{Research Question \#3:} ``What are some possible effects that {\small \textit{(i)}}~manual manipulation of the \textit{ML model's weights}, as well as {\small \textit{(ii)}}~various \textit{disturbances of the input} images -- both of which hypothetically resembling radiation-induced faults -- have on the performance of the ML model?''}

%% file: sections/5-results.tex
\section{Technical Implementation and Assessment}
\label{sec:assessment}
\noindent
As a constructive step-forward, our third contribution entails showcasing the effects of exemplary radiation-induced faults through original experiments---and releasing our developed source-code~\cite{ourRepo}. To this purpose, we first develop our baseline ML models for cloud detection~(§\ref{ssec:baseline}); then, we examine their robustness to manually introduced bit-flips~(§\ref{ssec:bitflip}); finally, we scrutinize their response when the faults affect the input test-data~(§\ref{ssec:perturbations}). \textit{Our evaluation is a proof-of-concept.}\footnote{\textbf{Problem:} we try to ``anticipate'' what effects radiation may have on the ML pipeline. \textit{Our anticipations are hypothetical}: there is no guarantee that any given sample be modified in the way we do it; and there is also no guarantee that ML model deployed on-board be impacted by natural radiation in the way we do it. We carry out our evaluation under the guidance of practitioners.} We carry out our experiments on a system mounting an Intel-Core i9 12900K with 32GB of RAM and an NVidia RTX3060Ti.

\subsection{Baseline: U-Net on Cloud-95 dataset}
\label{ssec:baseline}
\noindent
We align our assessment with our ``failed'' experiment (§\ref{ssec:negative}). 

\begin{figure}[!htbp]
    \centering
    \includegraphics[width=1\columnwidth]{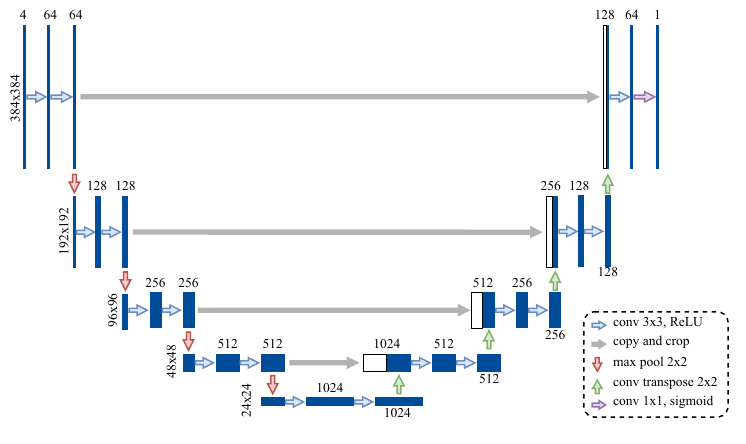}
    
    \caption{\textbf{Architecture of our U-Nets} --
    \textmd{\footnotesize The ML models used for our original experiments are drawn from the state of the art for cloud detection~\cite{mohajerani2019cloud}.}}
    \label{fig:unet}
    \vspace{-3mm}
\end{figure}

\textbf{Setup.}
We rely on the U-Net (shown in Fig.~\ref{fig:unet}): according to practitioners, this is the ML model they deploy on spacecraft for cloud detection. The dataset of choice is \dataset{95-Clouds}, from which we derive a train, \smabb{T}, and test, \smabb{E}, partition by applying an 80:20 split (common in related work~\cite{kalpoma2023deep}), which correspond to roughly 16k and 4k samples (after filtering-out some ``incorrect'' samples). Then, we extract a subset, \smabb{R}, of 50 samples from \smabb{E}: we will use this subset as a basis for our robustness assessment. This is necessary for a broad and comprehensive evaluation: in the real world, the effects of natural hazards are not deterministic (e.g., it is impossible to predict how any given sample may be affected by, say, some incorrect pixels). By considering such a small subset, we can gauge the impact of various types of faults (performing our assessment on the whole \smabb{E} would have required 80x longer runtime---amounting to several weeks of no-stop computing). Finally, to mitigate experimental bias, \textit{we develop three} U-Nets by keeping the same \smabb{T}, \smabb{E}, \smabb{R}, but by repeating the training from scratch with different randomly-initialized weights (repeating trials is fundamental in ML research). We set our the learning phase of our U-Nets to last for $\approx$3 hours, after which we stop (to avoid overfitting) and proceed with the assessment.

\vspace{-3mm}

\textbf{Results.} We report the baseline performance of our ML models (after learning) in Table~\ref{tab:baseline}, showing the accuracy, precision and recall of our U-Nets on \smabb{T}, \smabb{E} and \smabb{R}. We appreciate that, on average, our ML models achieve good performance on the test set (i.e., \smabb{E}), and comparable with state-of-the-art works (e.g.,~\cite{guo2021cloudet}). These findings confirm \textit{our baselines are a solid choice for our robustness assessment}. Our repository includes the detailed implementation~\cite{ourRepo}.

\vspace{-1mm}
\input{sections/tab_baseline}

\vspace{-2mm}

\subsection{Simulating radiation-induced bit-flips on the ML models}
\label{ssec:bitflip}
\noindent
An elusive property of naturally-occurring bit-flips is the impossibility of predicting \textit{which} and \textit{how many} bits will be flipped---i.e., each bit has the same probability of being flipped as any other bit. Hence, we consider various bit-flips.\footnote{\textbf{Background on bit-flips.} In simple terms, injecting bit-flips in an ML model means taking the bits representing the weights of the ML model and flipping them from a 0 to a 1, or from a 1 to a 0. In theory, changes to the ``first'' bits are likely to induce a greater impact to the final output of the ML model; whereas changes in the ``last'' bits are more likely to have a negligible effect. We also stress that a single flip in a ``high'' bit can be more impactful than multiple changes in ``lower'' bits.  For more detailed information, see:~\cite{rakin2019bit}.}

\begin{figure}[!htbp]
    \centering
    \includegraphics[width=1\columnwidth]{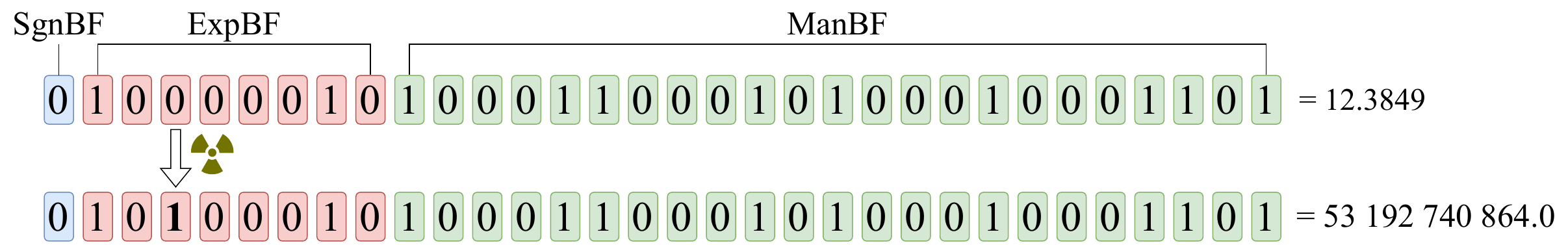}
    \vspace{-0.5cm}
    \caption{\textbf{Exemplary effects of a bit-flip} --
    \textmd{\footnotesize By flipping just a single \textit{exponential} bit (from a 0 to a 1), the value $12.38249$ changes to $53\,192\,740\,864.0$}.}
    \label{fig:bitflip_example}
    \vspace{-2mm}
\end{figure}

\textbf{Setup.} We take each (trained) U-Net, and we manually modify its learned weights and biases in three ways: first, a worst-case scenario (we provide an example in Fig.~\ref{fig:bitflip_example}) wherein we modify the bits of the \textit{exponent} (ExpBF); and two others wherein we modify the bits controlling the \textit{mantissa} (ManBF) or the \textit{sign} (SgnBF). Depending on the weight, ManBF can be more (or less) impactful than SgnBF. For a bias-free assessment, we consider bit-flips of one bit, which are chosen \textit{randomly} among the weights of each layer of our baseline models---each having 44 layers. We then run our ``faulty'' models again on the samples in the robustness set, \smabb{R}. We repeat this experiment 50 times to account for randomness.\footnote{Overall, we perform \scmath{3}(models)$\times$\scmath{44}(layers)$\times$\scmath{3}(types)$\times$\scmath{50}(repeats)=\scmath{19\,800} bit-flips. This entire assessment required 10h of non-stop computation.}

\vspace{-3mm}

\textbf{Results.} We report the results in Figs.~\ref{fig:bitflips_results}, showing the performance (y-axis), averaged across our 3 models, for each type of bit-flip which affects a given layer of our U-Net (x-axis); the lines represent the mean (blue), baseline (red) and the min/max range achieved during the 50 trials. Specifically, Fig~\ref{sfig:bitflips_acc} focuses on accuracy, Fig.~\ref{sfig:bitflips_pre} on precision and Fig.~\ref{sfig:bitflips_rec} on recall. Our repository includes instructions to perform additional assessments to further mitigate bias. By observing the accuracy (which is the most common metric) in Fig.~\ref{sfig:bitflips_acc}, we see that the majority of our single bit-flips have a negligible impact. Notably, however, for ExpBF the performance of our ``faulty'' models decreases the most when the bit-flip affects the outer layers of the U-Net; whereas only the first layers are affected in SgnBF---albeit at a much lower degree than for ExpBF. Intriguingly, ManBF may increase the performance.

\textbf{Analysis.}
We find it instructive to compare this experiment with the one by Haser and Förstner~\cite{haser2022Reliability}. Specifically, the evaluation in~\cite{haser2022Reliability} seeks to assess the robustness of neural networks against bit-flips on the \dataset{MNIST} dataset (we posit that this setting is highly unrepresentative of a realistic space environment). Nevertheless, the results in~\cite{haser2022Reliability} show that the accuracy of the ML model drops to that of a coin-toss by flipping 700 bits; and that, e.g., SgnBF affecting over 50\% of the layers also leads to an unusable model. According to our practitioners, these findings, while useful, represent ``unlikely'' circumstances: even in space missions, such high corruptions are unlikely. This is why we opted for single bit-flips, which are more likely to occur and also more subtle. Moreover, Haser and Förstner~\cite{haser2022Reliability} conclude that ``if a corruption occurs in later parts its impact is more harmful then in early layers'' (sic). This is in contrast with our findings: e.g., for ExpBF, bit-flips affecting last \textit{and} first layers are more impactful than on middle layers (result confirmed with a statistical t-test: \smamath{p\approx0}). 

\vspace{-3mm}
\input{sections/fig_bitflips}

\subsection{Effects of radiation-induced perturbations on image data}
\label{ssec:perturbations}
\noindent
Images can be affected by natural hazards in many ways (see Fig.~\ref{fig:disturbances} in §\ref{sssec:effect_data}). Here, we only consider those disturbances that are more common in space---according to practitioners.

\begin{figure}[!htbp]
    \centering
    \includegraphics[width=1\columnwidth]{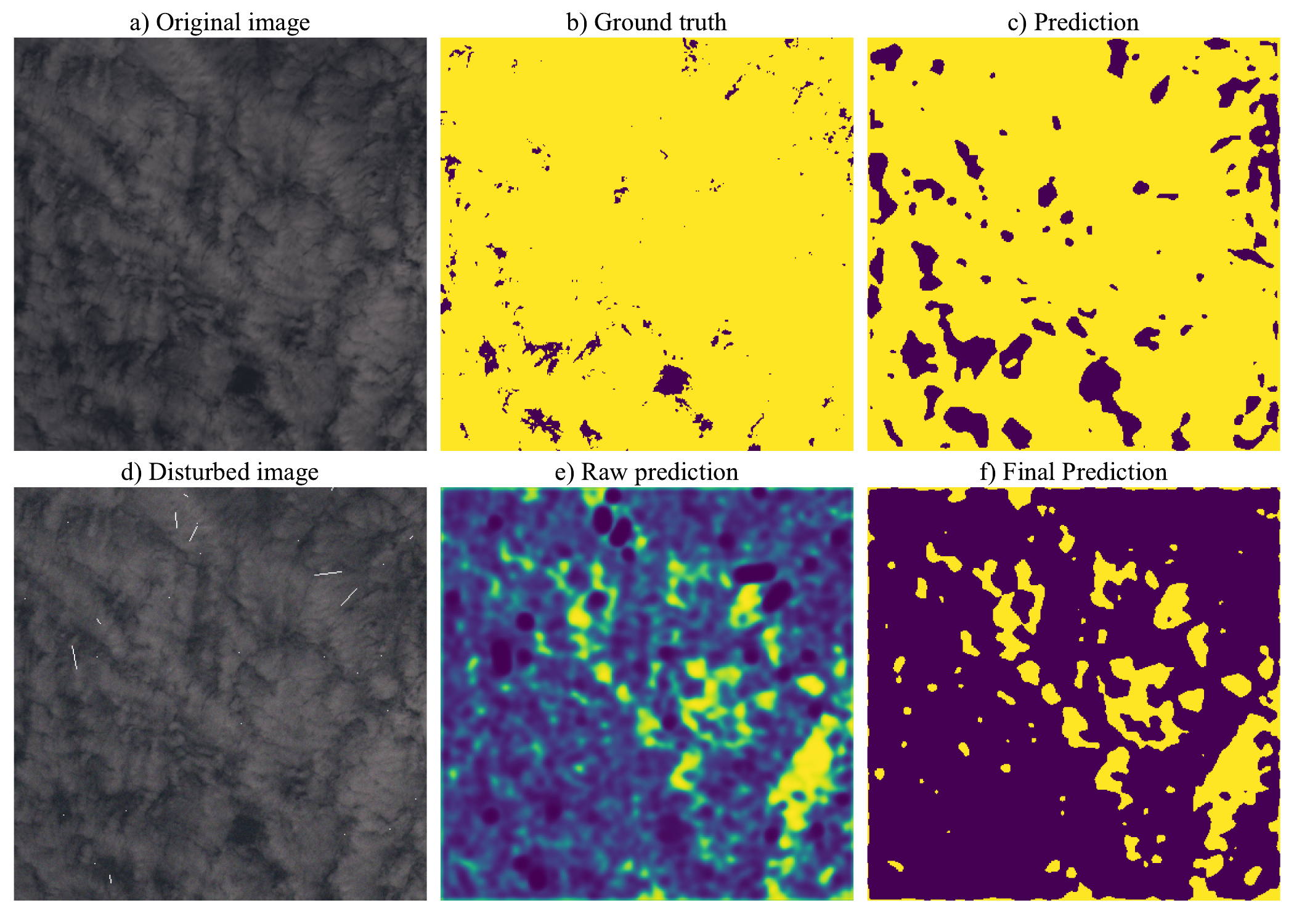}
    \vspace{-0.6cm}
    \caption{\textbf{Exemplary effect of disturbances} --
    \textmd{\footnotesize We show how the effects of introducing common disturbances in spacecrafts (hot pixels, radiation streaks, dark currents) to the input images during the ML model's inference.}}
    \label{fig:radiation_streak}
    \vspace{-3mm}
\end{figure}

\textbf{Setup.} We consider three types of disturbances: \textit{hot} pixels, which involve having some pixels of an image be ``brighter'' w.r.t. their originals; \textit{dark currents}, which yield a fixed-pattern noise in an image; and \textit{radiation streak}, which involve having multiple pixels in succession to (incorrectly) have the same value. According to practitioners, these disturbances are very common in the images captured by operational spacecraft. We provide an exemplary illustration of wherein we combine all these disturbances (alongside their effects on the ML model) in Fig.~\ref{fig:radiation_streak}.
For each of the 50 samples in \smabb{R}, we apply each disturbance at various noise levels (e.g., we progressively increase the number of faulty pixels in an image); the areas affected are chosen randomly. Our repository provides the detailed implementation~\cite{ourRepo}. After applying these manipulations, we run our (baseline) U-Nets again on the perturbed images.

\begin{figure}[!htbp]
    \centering
    \includegraphics[width=1\columnwidth]{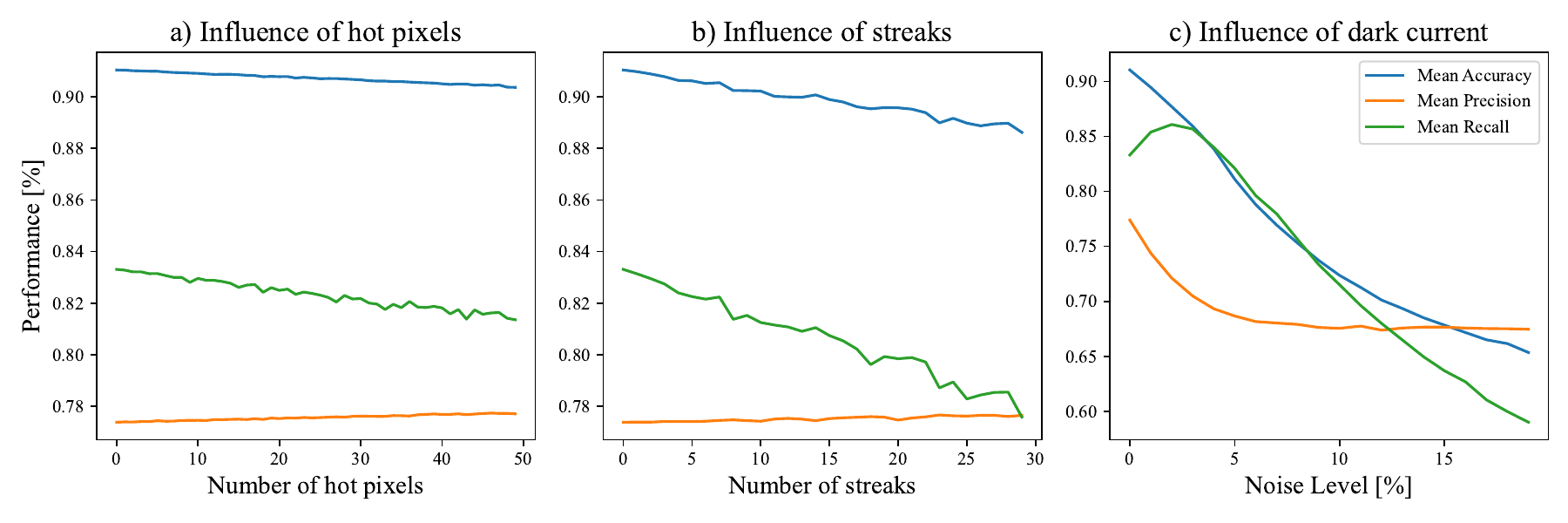}
    \caption{\textbf{Effects of image disturbances} --
    \textmd{\footnotesize We compute the performance (averaged across our three U-Net) on the images in \scbb{R} after applying our disturbances at increasing noise levels.}}
    \label{fig:disturbances-results}
    \vspace{-3mm}
\end{figure}

\textbf{Results.} We report the results in Fig.~\ref{fig:disturbances-results}, showing the average accuracy, precision and recall (y-axis) achieved by our three U-Nets at varying intensity (x-axis) of each disturbance applied to all samples in \smabb{R}. Our repository includes the code to run this evaluation on different images, as well as to repeat these experiments at different noise levels and/or more times. From Fig.~\ref{fig:disturbances-results}, we observe that the performance decreases for increasing noise levels, especially \textit{dark currents} and \textit{streaks}. 

\textbf{Analysis.} We find it interesting, however, that \textit{hot pixels} have a relatively mild effect. This result contrasts the impact of targeted ``one-pixel'' attacks that are sometimes considered in related papers on adversarial ML robustness~\cite{su2019one}. From a security standpoint, this result underscores that radiation-induced faults must be treated differently than ``adversarial examples'', whose countermeasures (e.g., adversarial training) induce a decrease in the baseline performance of an ML model~\cite{apruzzese2023real} (which may not be justified in space-contexts).

\vspace{-3mm}

%% file: sections/tab_baseline.tex
\newcommand\res[1]{\scriptsize {$#1$}}

\begin{table}[!htbp]
    \centering
    \caption{\textbf{Baseline --}
    \textmd{\footnotesize We train three U-Nets (on \scbb{T} for 3h). Then, we compute accuracy (\textit{Acc}), precision (\textit{Pre}) and recall (\textit{Rec}) on our selected subsets (\scbb{T}, \scbb{E}, \scbb{R}) of \dataset{95-Cloud}. Our results (on \scbb{E}) match those of prior works (e.g.,~\cite{guo2021cloudet}).}}
    \label{tab:baseline}
    \vspace{-2mm}
    \resizebox{\columnwidth}{!}{
        \begin{tabular}{c||c|c|c?c|c|c?c|c|c}
            \toprule
            \multirow{2}{*}{\textbf{Model}} & \multicolumn{3}{c?}{\textbf{Training Set (\scbb{T})}} & \multicolumn{3}{c?}{\textbf{Test Set (\scbb{E})}} & \multicolumn{3}{c}{\textbf{Robustness Set (\scbb{R})}} \\ \cline{2-10}
            
            & \textit{Acc} & \textit{Pre} & \textit{Rec} & \textit{Acc} & \textit{Pre} & \textit{Rec} & \textit{Acc} & \textit{Pre} & \textit{Rec} \\ 
            \midrule
            
            U-Net \#1 & \res{90.1} & \res{92.9} & \res{82.9} & \res{89.9} & \res{92.5} & \res{81.7} & \res{91.2} & \res{93.5} & \res{84.1} \\
            U-Net \#2 & \res{89.1} & \res{83.3} & \res{92.4} & \res{88.9} & \res{82.8} & \res{91.6} & \res{91.6} & \res{87.7} & \res{93.0} \\
            U-Net \#3 & \res{88.1} & \res{81.5} & \res{92.7} & \res{87.8} & \res{81.0} & \res{92.0} & \res{90.4} & \res{84.7} & \res{94.8} \\
            \midrule
            average & \res{89.1} & \res{85.9} & \res{89.3} & \res{88.9} & \res{85.4} & \res{88.4} & \res{91.1} & \res{88.6} & \res{90.7} \\
            \bottomrule
        \end{tabular}
    }
\end{table}

%% file: sections/fig_bitflips.tex
\begin{figure}[!htbp]
    \centering
    \begin{subfigure}[!htbp]{1\columnwidth}
        \centering
        \includegraphics[width=1\columnwidth]{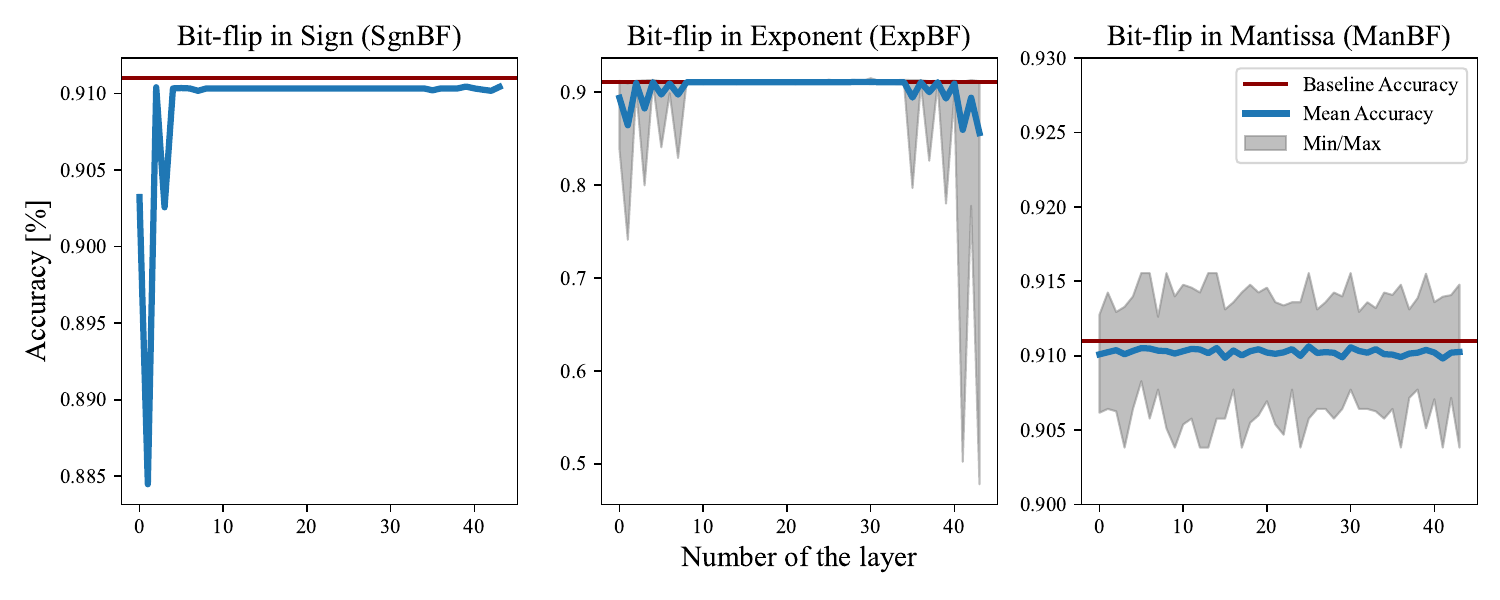}
    \vspace{-0.6cm}
        \caption{Accuracy.}
        \label{sfig:bitflips_acc}
    \end{subfigure}
    \begin{subfigure}[!htbp]{1\columnwidth}
        \centering
        \includegraphics[width=1\columnwidth]{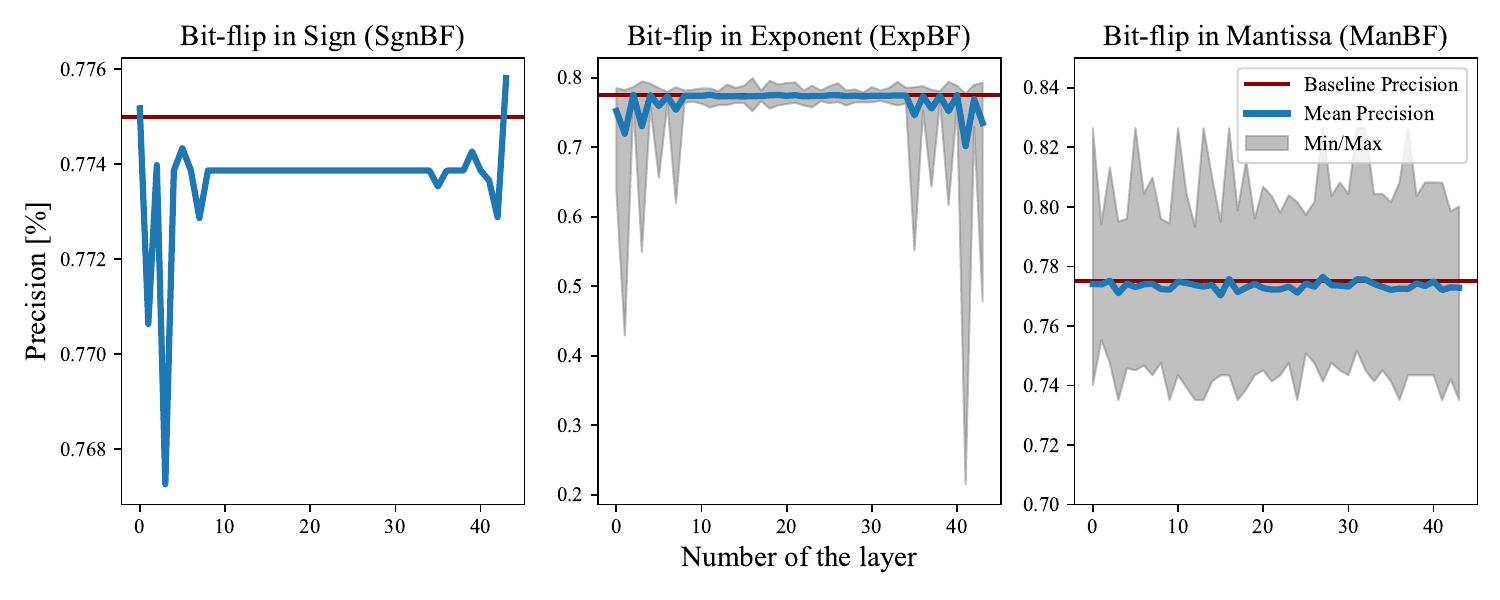}
    \vspace{-0.6cm}
        \caption{Precision.}
        \label{sfig:bitflips_pre}
    \end{subfigure}
    \begin{subfigure}[!htbp]{1\columnwidth}
        \centering
        \includegraphics[width=1\columnwidth]{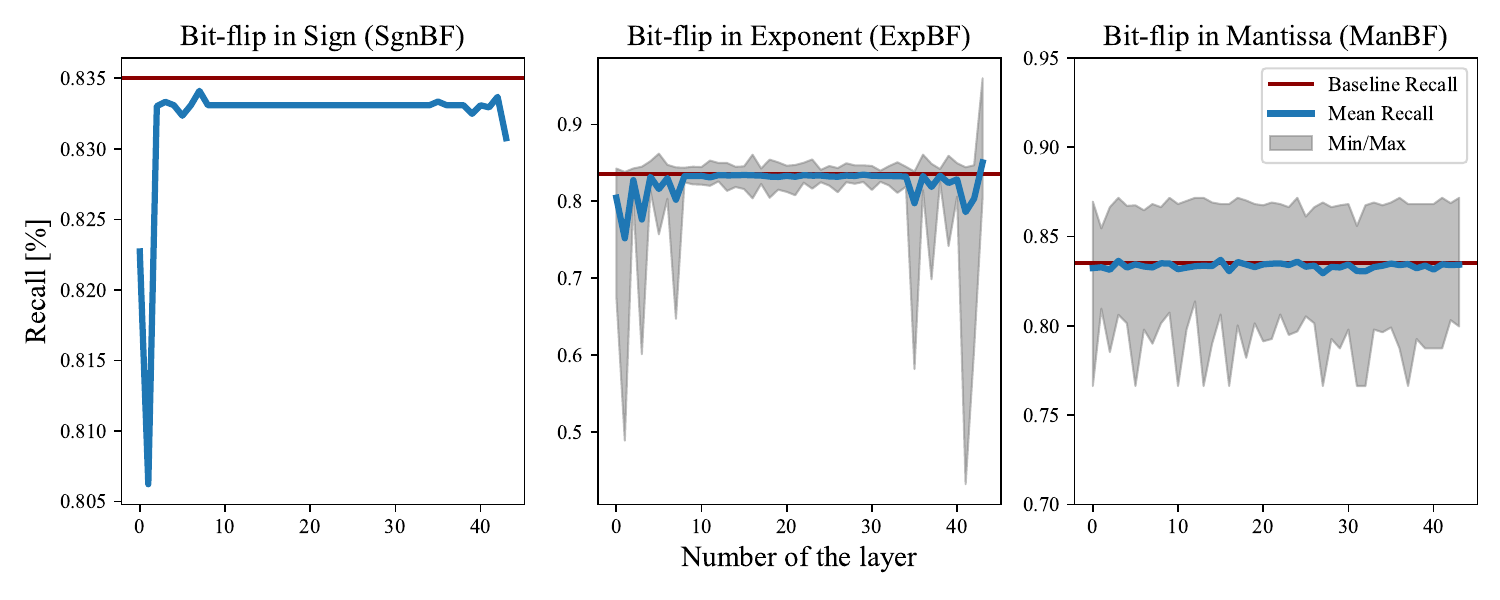}
    \vspace{-0.6cm}
        \caption{Recall.}
        \label{sfig:bitflips_rec}
    \end{subfigure}
    \vspace{-2mm}
    \caption{\textbf{Effects of bit-flips} --
    \textmd{\footnotesize We measure the performance (averaged across 50 trials) of our bit-flips on each possible layer (out of 44) of our 3 U-Nets.}}
    \label{fig:bitflips_results}
    \vspace{-3mm}
\end{figure}

%% file: sections/6-discussion.tex
\section{Discussion and Implications}
\label{sec:discussion}
\noindent
We now critically review our study, identifying limitations and clarifying how our findings are useful to related research.

\subsection{Disclaimers (and Alternative Formulations)}
\label{ssec:limitations}
\noindent
We acknowledge that our work may have limitations. Let us reflectively scrutinize each of our contributions.

\textbf{Literature Review (§\ref{sec:sota}).} The papers included in Table~\ref{tab:sota} have been chosen by two authors who queried popular scientific resources for publications that envisioned applications of ML for on-board spacecraft. For instance, papers proposing an ML method for cloud detection in images but which do not specifically aim (or mention) at on-board satellite deployment, such as~\cite{abderrahmane2022spleat}, were omitted from our analysis. We did so because our goal was assessing the the extent to which prior work on ML for space accounts for its natural hazards. Furthermore, we did not perform the forward/backward search ad infinitum (we inspected 362 documents just from 2 papers), so we did not include papers that matched our inclusion criteria if they were cited (or cited by) in a longer chain. For instance, we did not include the paper by Sabogal et al.~\cite{sabogal2019recon} (which carries out a detailed assessment on dedicated hardware---including a real radiation-beam!). We also did not consider several works (\cite{veyette2022ai, esposito2019hyperscout, esposito2019highly,chavier2023deploying,ruszczak2023machine,fanizza2022transfer,macdonald2022enabling}) because they were behind a paywall we could not overcome (even through our own institutions).
Nonetheless, our findings revealed that abundant prior work on ML in space poorly accounted for natural hazards, and that few release their resources publicly. However, we acknowledge that there is a growing body of research (notably, from the University of Pittsburgh~\cite{wilson2017hybrid,sabogal2021methodology,kain2020evaluating}) that is investigating such phenomena. Our work can inspire future endeavours to better account for the issues that affect deployment of ML in spacecraft---and openly release corresponding toolkits to the research community.

\textbf{Tool Analysis (§\ref{sec:tools}).} Our survey of state-of-the-art technologies for ML-based experiments in space contexts was limited to \textit{publicly available} solutions, which have been found by two researchers supported by practitioners. We are aware that we may have overlooked, e.g., some repositories, and that some closed-source resources (potentially available upon request) may exist which may allow a researcher to simulate realistic testbeds (e.g.,~\cite{luza2021emulating}). Furthermore, in our analysis we excluded PyTorchFI (used, e.g., in~\cite{shi2023automated}---whose code is not released) because it was deemed impractical by practitioners. Finally,  our low-level experiments (for \fmwrk{NVBitFI} and \fmwrk{LLTFI}) have been carried out by individuals who, despite having plenty of experience (over 5 years) in software development and/or ML, may have overlooked some details.\footnote{\textbf{Ethics:} we contacted the maintainers of the \ftfmwrk{NVBitFI} and \ftfmwrk{LLFTI} repositories, informing them of the issues we encountered while using these tools.} Nonetheless, reproducibility issues are---unfortunately---common in the ML domain (e.g.,~\cite{olszewski2023get}). To fix this, and also for complete transparency, we disclose our code and procedures~\cite{ourRepo}.

\textbf{Proof-of-Concept Experiments (§\ref{sec:assessment}).} Our assessment is an effort to spearhead novel research that accounts for radiation-induced faults in space-related evaluations---with the ultimate goal of finding ways to mitigate this real problem. Our results entail the application of well-known methods (U-Net) on benchmark datasets (\dataset{95-Cloud}) and by simulating faults (bit-flips or image disturbances). However, we cannot claim our testbed has 100\% fidelity with a real spacecraft---albeit the practitioners that we inquired confirmed our workflow to represent a realistic pipeline. Moreover, we do not claim novelty in the methods we use---albeit we do claim our findings to be original (especially in light of the conclusions in~\cite{haser2022Reliability}). Finally, there are potentially infinite ways to inject the faults we considered. We account for this by releasing our source-code~\cite{ourRepo}: the interested researcher can replicate our experiments, or carry out new evaluations by considering different faults, models, or datasets. 

\begin{cooltextbox}
\textsc{\textbf{Takeaways.}} Our technical contributions are addressed at ML enthusiasts  who want to develop of space-tolerant ML software \textit{without} relying on specialized hardware.\vspace{-2mm}\footnote{§\ref{sec:sota} and §\ref{sec:tools} showed that existing solutions have practical limitations.}
\rule{0.2\textwidth}{0.1pt} 
\vspace{-3mm}
\end{cooltextbox}

\subsection{Implications for Research (and for Practice)}
\label{ssec:implications}
\noindent
We focus on deployment of ML in spacecraft---a setting that partially overlaps with, e.g., aerial vehicles and communications~\cite{ding2023get,strohmeier2018k,dick2023research} or GPS~\cite{xue2020deepsim}. We focus on solutions at the software-level: developing radiation-tolerant hardware (e.g.,~\cite{martin2022radiation}) is an orthogonal research field.
Our contributions can be leveraged by a wide audience: we discuss 3 groups.

\textbf{Power Consumption.} Many papers on space missions consider the relationship between ML and power consumption~\cite{perryman2023evaluation,franconi2023comparison,kocik2021space}. For instance,~\cite{kocik2021space} propose to use ML to forecast energy requirements, whereas~\cite{perryman2023evaluation} assess the power consumption of equipment which can also empower ML models. Our contributions can be helpful to this research area: indeed, they would inspire ML researchers ``on Earth'' to gauge the accuracy applications of ML similar to those envisioned in~\cite{kocik2021space} by simulating certain faults, and then devising appropriate ML-specific hardening methods. Alternatively, since radiation can degrade the performance of ML also \textit{during training}~\cite{dong2023one}, it would be intriguing to assess the energy expenditure of ML models under the impact of radiation-induced bitflips: such an analysis would be useful for those applications that envision on-board ML training~\cite{labreche2022ops}.

\textbf{Space Security.}
We focus on the impact of naturally-induced faults on ML models---a problem that falls in the ``robustness'' research domain. Our findings are hence closely related to cybersecurity~\cite{lin2023clextract}. E.g., papers on ``adversarial ML'' (e.g.,~\cite{apruzzese2023real}) envisage ``data perturbations'', which are strongly connected to the perturbations that we injected for our image experiments (§\ref{ssec:disturbance}). Notably, some works envision ``adversarial attacks'' against object detectors meant for space deployment~\cite{du2022adversarial}, but \textit{do not account for radiation}: it would be intriguing to study the effects of such ``adversarial perturbations'' if combined to the effects of naturally-occuring perturbations. Finally, satellites are now targeted by real attackers~\cite{willbold2023space}, and must hence be protected~\cite{fischer2015finalizing,adalier2020efficient} against sophisticated cyberthreats (e.g.,~\cite{falco2023wannafly, giuliari2021icarus}). 
Our findings (and resources) can open research avenues considering, e.g., the effects of radiation-induced faults on ML models for on-board network security (e.g.,~\cite{smailes2023watch}).

\textbf{Developers and Practitioners.}
Among the main take-home messages of our paper is that there is a lack of publicly available resources for realistic assessments of ML methods in space contexts. Indeed, our work would not have been possible without the guidance of practitioners, who provided us with valuable information on how ML is deployed in spacecraft. Put simply, we argue that {\small \textit{(i)}}~the shortage of ``plug-and-play'' (and open-source) tools for radiation-induced faults, paired with {\small \textit{(ii)}}~the prohibitive costs to setup a representative testbed for space-related assessments is a substantial barrier for research breakthroughs. For instance, the ML community flourished thanks to the open release of code and data; yet, the results claimed by research papers on generic ML have questionable value for deployment of ML on-board satellites, due to the ``naturally adversarial'' environment of spacecraft---of which we know very little about from the ML perspective. Hence, \textit{we advocate developers at all levels to prioritize publicly accessible and easy-to-use tools for space experiments} (and, in particular, for robustness assessments of ML methods).

\vspace{1mm}

{\setstretch{0.8}
\textbox{{\small \textbf{Call to Action.} To quote Crum et al.~\cite{crum2022nasa}:
``Progress in this field depends on many stakeholders working together efficiently; not only scientists from the astronomy and physics, ground- and space-based communities, but also engineers, software developers, data, and communication scientists, and more'' and ``to help development [...] an open simulation and test environment is needed''.}}}

%% file: sections/7-conclusions.tex
\section{Conclusions and Future Work}
\label{sec:conclusions}
\noindent
We seek to improve the robustness of ML models deployed on-board spacecraft to radiation-induced faults. 
We reveal that {\small \textit{(i)}}~prior research poorly accounted for the natural hazards that may impact the performance of ML in space; and that {\small \textit{(ii)}}~current open-source technologies are poorly suited to examine this problem from the perspective of an ML researcher. To improve the current situation, we pair-up with practitioners and carry out proof-of-concept experiments highlighting the effects of some radiation-induced faults on state-of-the-art ML methods for cloud detection. Moreover, to foster development of future efforts focusing on radiation-tolerant ML components, we release all our tools and data in a dedicated repository~\cite{ourRepo}. We also include a \textbf{demonstrative 2-minutes \href{https://github.com/langekevin/mlspace_robustness/blob/main/assets/video/demonstrative_video_720p.mp4}{video}}, showcasing the simplicity of using our resources. 

As intriguing avenues for research that can be built upon this work, we suggest: the assessment of radiation-induced faults \textit{at training-time} (we only considered effects at the inference-stage), since ML models should be periodically updated and re-trained; and the consideration of ML models that analyse data \textit{different from images} (e.g., power or network data).

\let\thefootnote\relax\footnotetext{\textbf{\textsc{Acknowledgement.}} We would like to thank the anonymous reviewers of the SpaceSec workshop for the invaluable feedback. We also thank the Hilti Corporation for funding part of this research.}